\documentclass[letterpaper, 10 pt, conference]{ieeeconf}  

\IEEEoverridecommandlockouts                              
\usepackage{shortcuts}

\title{\LARGE \bf
Lateral Ego-Vehicle Control without Supervision using Point Clouds\thanks{This work was supported by the Munich Center for Machine Learning}
}

\author{Florian Müller*, Qadeer Khan* and Daniel Cremers
\thanks{* These authors contributed equally}}%

\affil{Computer Vision and Artificial Intelligence Group \\ Technical University of Munich, Germany\\ {\{f.r.mueller, qadeer.khan, cremers\}@tum.de} }

\begin{document}

\maketitle

\begin{abstract}
  Existing vision based supervised approaches to lateral vehicle control are capable of directly mapping RGB images to the appropriate steering commands. However, they are prone to suffering from inadequate robustness in real world scenarios due to a lack of failure cases in the training data. In this paper, a framework for training a more robust and scalable model for lateral vehicle control is proposed. The framework only requires an unlabeled sequence of RGB images. The trained model takes a point cloud as input and predicts the lateral offset to a subsequent frame from which the steering angle is inferred. The frame poses are in turn obtained from visual odometry. The point cloud is conceived by projecting dense depth maps into 3D.  An arbitrary number of additional trajectories from this point cloud can be generated during training. This  is to increase the robustness of the model. Online experiments show that the performance of our method is superior to that of the supervised model.   
\end{abstract}

\section{Introduction}\label{chapter:introduction}

In recent years, deep learning approaches have shown a promising trend in the context of lateral sensorimotor control \cite{toromanoff2018end,chen2015deepdriving,Pomerleau1988ALVINNAA}. The trained network can directly map input data to the steering commands \cite{7995975,bojarski2016end}.  Labeled training data is usually acquired by recording the raw sensory input and the corresponding steering commands executed by an expert driver traversing a reference trajectory. One of the main challenges of this approach is the lack of failure cases in the training data, caused by the driver's obligation to follow traffic rules and to remain within its own driving lane. Without failure cases in the training data, the model has no way of learning to recover from a divergence from the reference trajectory \cite{ross2011reduction}. This is a common issue with deep learning that models tend to fail at inference time when encountering  images that are out-of-distribution from the training set \cite{8793742}. Previous works have attempted to solve this problem by generating training images with lateral displacement and adjusted steering label \cite{bojarski2016end,hubschneider2017adding,toromanoff2018end}. However, due to limitations in the maximum lateral offset they can generate, the learned driving policy tends to be not robust enough \cite{codevilla2017endtoend}. 

In contrast to supervised approaches, Reinforcement Learning (RL) can be used to learn a driving policy for lateral control \cite{8851766,8686348}. RL does not require explicit data-label pairs for training. Rather, the model learns a suitable policy by randomly exploring the environment in a hit and trial method using a pre-defined reward function \cite{sutton2018reinforcement}. However, in the context of self-driving, random exploration of the driving environment is not a feasible solution as it may involve dangerous traffic violation, potentially causing crashes. This is why \cite{8793742}, have an expert driver that assumes control whenever the car starts to deviates off-course. 

\begin{figure*}[ht]
    \centering
    \includegraphics[width=1\textwidth]{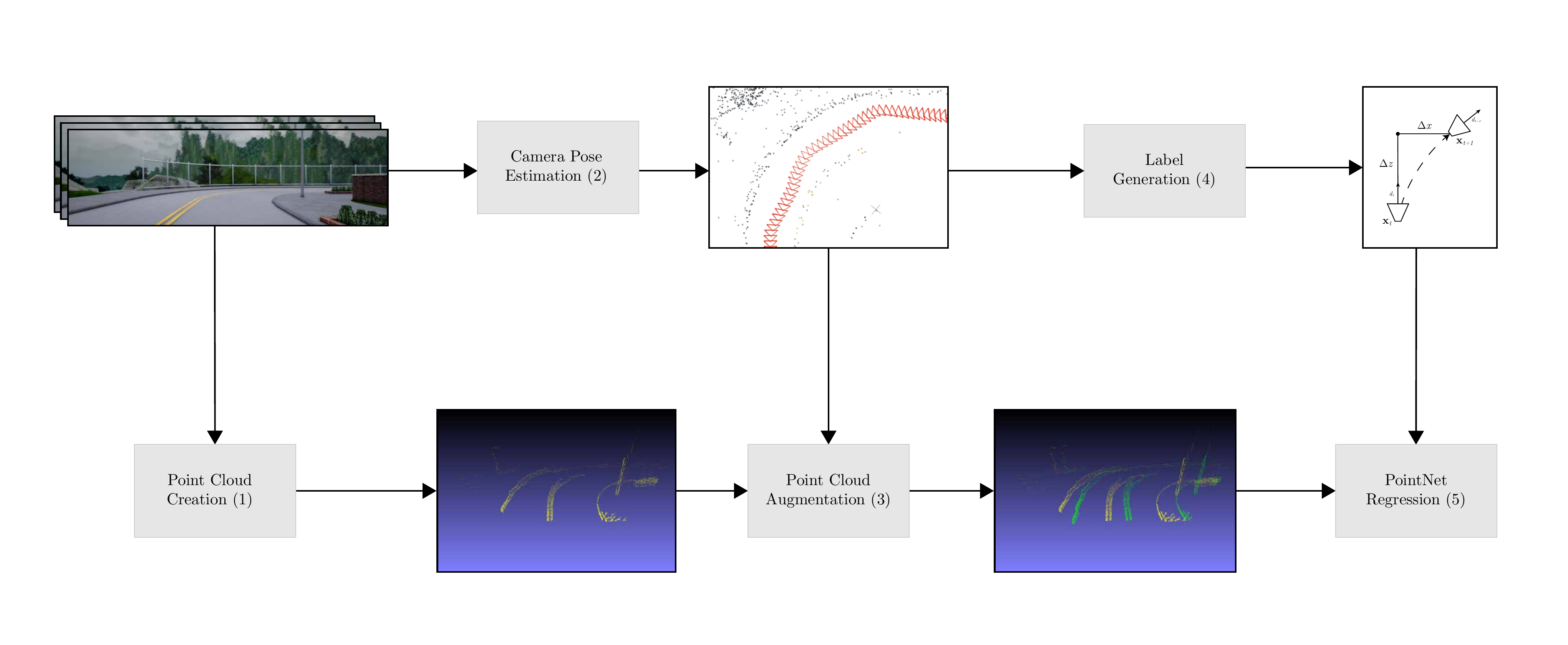}

    \caption{This figure describes the high level overview of the proposed framework (1) Point clouds are generated from RGB images (Figure~\ref{fig:pointcloud_pipeline}, Section~\ref{subsection:monodepth}). (2) The camera pose of each frame is estimated using visual odometry (Section~\ref{subsection:pose_estimation}).(3) New camera frames are generated by aligning, shifting and cropping point clouds from existing frames (Section~\ref{subsection:data_generation}). (4) Labels representing the offset of the next frame in lateral direction are generated from the reconstructed camera poses (Section~\ref{subsection:label_generation}). (5) The resulting point clouds and labels are used to train a deep learning model to predict the lateral offset of the next frame given a point cloud (Section~\ref{subsection:pointnet}).}
    \label{fig:general_pipeline}
\end{figure*}

As depicted in Figure \ref{fig:general_pipeline}, we propose a scalable framework that neither requires supervised labels nor needs data collected in violation of traffic rules. The model learns to predict the lateral offset to subsequent frames from an unlabeled sequence of RGB images, captured with a single front facing camera. Internally, the input images are converted to 3D point clouds, which enables the generation of an arbitrary number of realistic synthetic trajectories parallel to the reference trajectory. The labels for each frame are generated from reconstructed camera poses using visual odometry. This removes the requirement of labeled training data. 
\\
The primary contributions of our work are summarized below:
\begin{enumerate}
    \item We demonstrate how a model for lateral vehicle control can be trained from only an unlabeled sequence of images. 
    \item We show how generating additional training data  leads to enhanced robustness of the model at inference time.  
    
\end{enumerate}

\section{Method}\label{chapter:method}

In this section the individual components of our framework depicted in Figure \ref{fig:general_pipeline} are described in further detail. The framework trains a deep learning model for the task of lateral ego-vehicle control from an unlabeled sequence of RGB images.  

A high-level overview of the individual components is summarized below

\begin{enumerate}
    \item \textbf{Point Cloud Generation:} Using a self-supervised training paradigm, depth maps are predicted from the RGB images, which are then used to generate 3D point clouds.
    \item \textbf{Camera Pose Estimation:} Using a general-purpose visual odometry pipeline, camera poses are estimated for each frame.
    \item \textbf{Point Cloud Augmentation:} By aligning, shifting and cropping the point clouds, new point clouds are synthesized, which simulate additional trajectories.
    \item  \textbf{Label Generation:} Using the camera poses, the lateral offset to a subsequent frame is calculated for each frame. This serves as a target label to train the model.
    \item \textbf{Model Training:} A deep learning model is trained to predict the lateral offset to the next frame given a point cloud as input.

\end{enumerate}
These components are described in detail in the following subsections.
\subsection{Point Cloud Generation}\label{subsection:monodepth}

\begin{figure*}[h]
    \centering
    \includegraphics[width=0.9\textwidth]{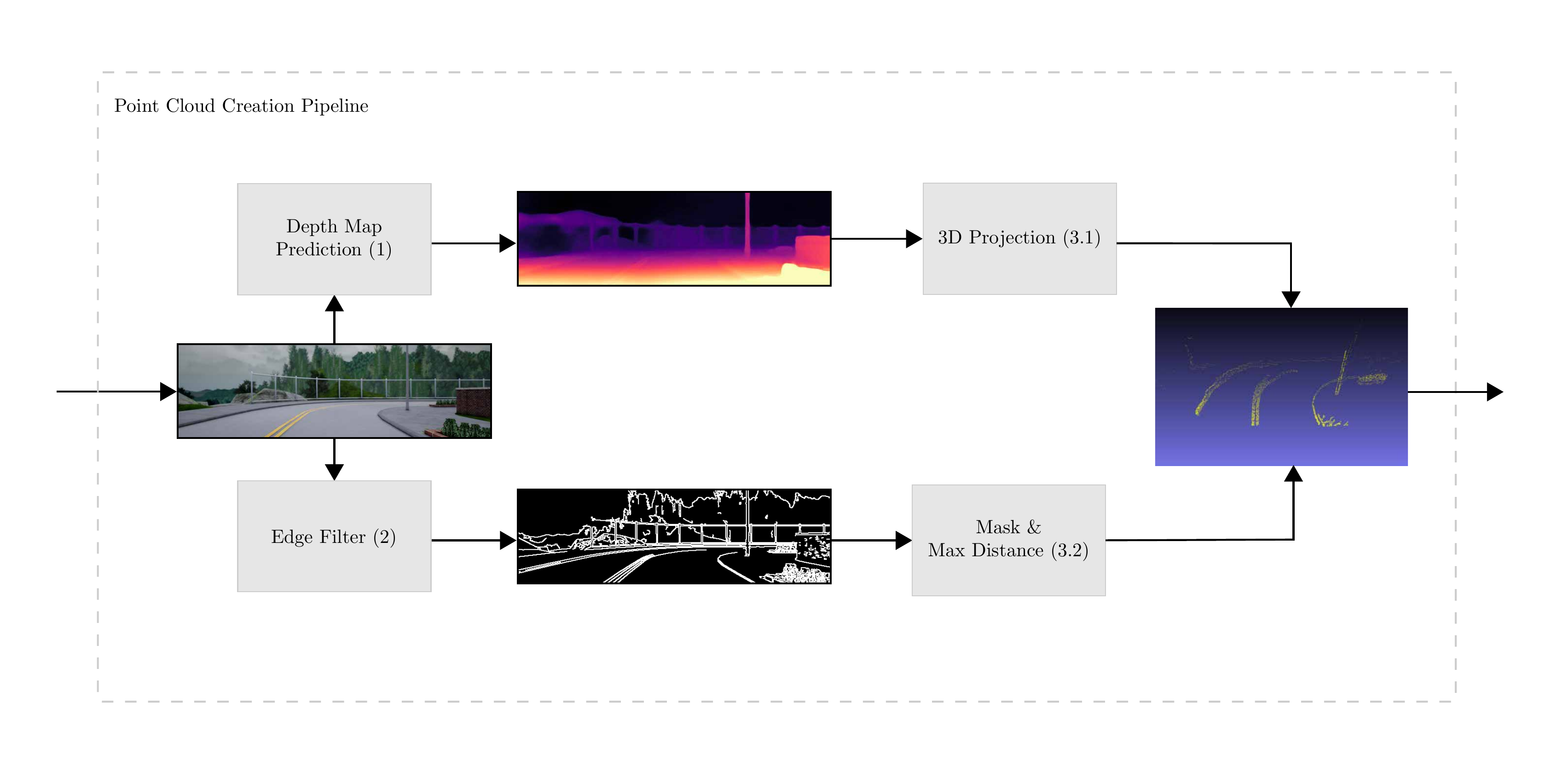}
    \caption{The figure describes the high level overview of the point cloud creation pipeline. An (inverse) depth map is predicted from an input RGB image, which in turn is projected to a 3D point cloud. Additionally an edge filter is applied to filter the point cloud by removing points beyond a certain threshold distance}
    \label{fig:pointcloud_pipeline}
\end{figure*}

Note that the point cloud for a corresponding RGB image is needed for 2 purposes: 
\begin{enumerate}
 \item As an input to the model for predicting the steering angle for lateral vehicle control
 \item It is used for synthesizing additional training trajectories. Hence, the data collection along trajectories which would otherwise violate traffic rules can be avoided. 
\end{enumerate}
Figure \ref{fig:pointcloud_pipeline} shows an example of  a point cloud projected into 3D using the dense depth map of a corresponding RGB image. The depth map is obtained from a depth  estimation  network  which takes an RGB image as input and can be trained in an entirely self-supervised manner \cite{monodepth2}. The training only requires a monocular sequence of RGB images. Note that the depth produced is normalized. Hence, calibration needs to be done to find the appropriate factor to scale the normalized depth map to the world scale.

 In the point clouds generated from dense depth maps, it may be hard to recognize relevant high-level features like road markings, traffic poles etc. These high level features tend to be important for the model to take the appropriate steering decisions \cite{8968451}. Moreover, having redundant points in the point cloud which do not yield useful information for the vehicle control model would impose an unnecessary computational burden.  
To counteract both issues, an edge filter \cite{4767851} is applied on the RGB image used to predict the depth map. The resulting point cloud is then filtered to only include points on the detected edges. The edges can further be dilated by a small amount to prevent losing finer details. Additionally, points beyond a certain distance from the camera are discarded. This increases the concentration of points in the relevant areas closer to the camera. Points closer to the camera are more important for the control model for immediate decision making.

\subsection{Camera Pose Estimation:}\label{subsection:pose_estimation}

\begin{figure}[h]
    \centering
    \includegraphics[width=0.5\textwidth]{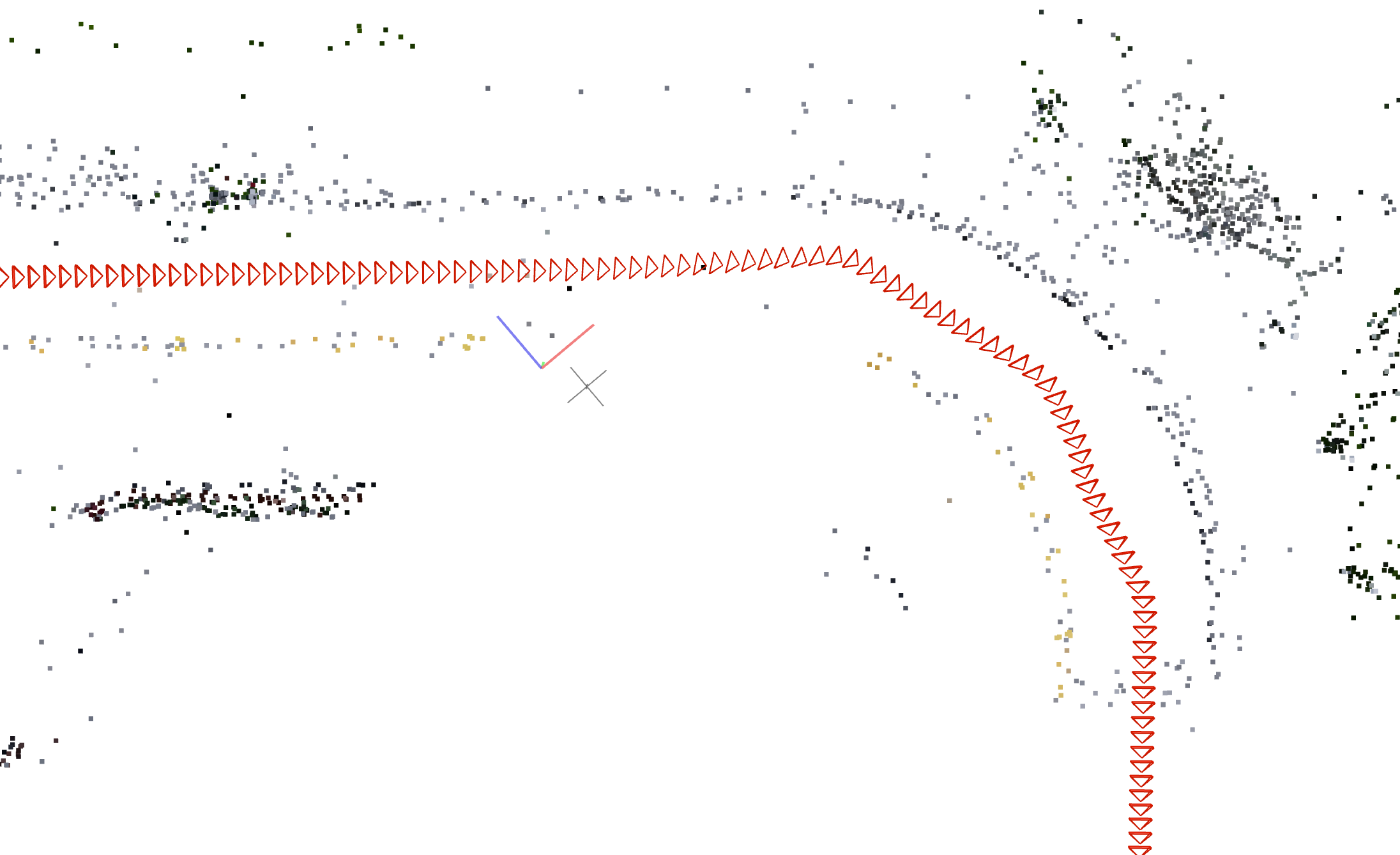}
    \caption{ Using visual odometry, the camera poses and a sparse 3D representation is reconstructed simultaneously from the sequence of RGB images. The figure shows a birds eye view on the reconstructed trajectory for a small section of the road. Each red triangle represents the reconstructed pose of one camera frame in the image sequence. The dots represent a sparse reconstruction of the 3D scene and are obtained by matching features extracted from the images in multiple frames.}
    \label{fig:reconstruction}
\end{figure}

The monocular sequence of RGB images contains no explicit information about the camera poses. The poses serve two purposes:
\begin{enumerate}
 \item To determine the target labels when training the network for lateral ego-vehicle control. The input to this network is the filtered point cloud. 
 \item For aligning point clouds when  additional trajectories are generated.
\end{enumerate}
The camera poses in the global frame of reference can be obtained  from the image sequence using a general-purpose visual odometry pipeline such as \cite{murAcceptedTRO2015,schoenberger2016sfm,schoenberger2016mvs}. They produce the 6DoF camera pose and a sparse 3D representation of the scene as depicted in Figure \ref{fig:reconstruction}. The camera pose can be expressed as a transformation matrix belonging to the special Euclidean group SE(3) representing a rigid body motion.

The path traversed by a sequence of these camera poses is referred to as the \emph{reference trajectory}. However, note that the camera poses obtained from a monocular sequence of images does not necessarily reflect the actual scale.  As was the case in Section \ref{subsection:monodepth}, calibration is done to determine the appropriate scaling factor.
Note that the  3D reconstruction of the scene obtained by running visual odometry is too sparse to be used as an input to the vehicle control model. This is why we have used the approach described in Section \ref{subsection:monodepth} that is capable of generating dense depth maps, while being trained in a completely self-supervised manner.

\subsection{Point Cloud Augmentation}\label{subsection:data_generation}
\begin{figure}[h]
    \begin{subfigure}{0.5\textwidth}
        \centering
        \includegraphics[width=0.7\linewidth]{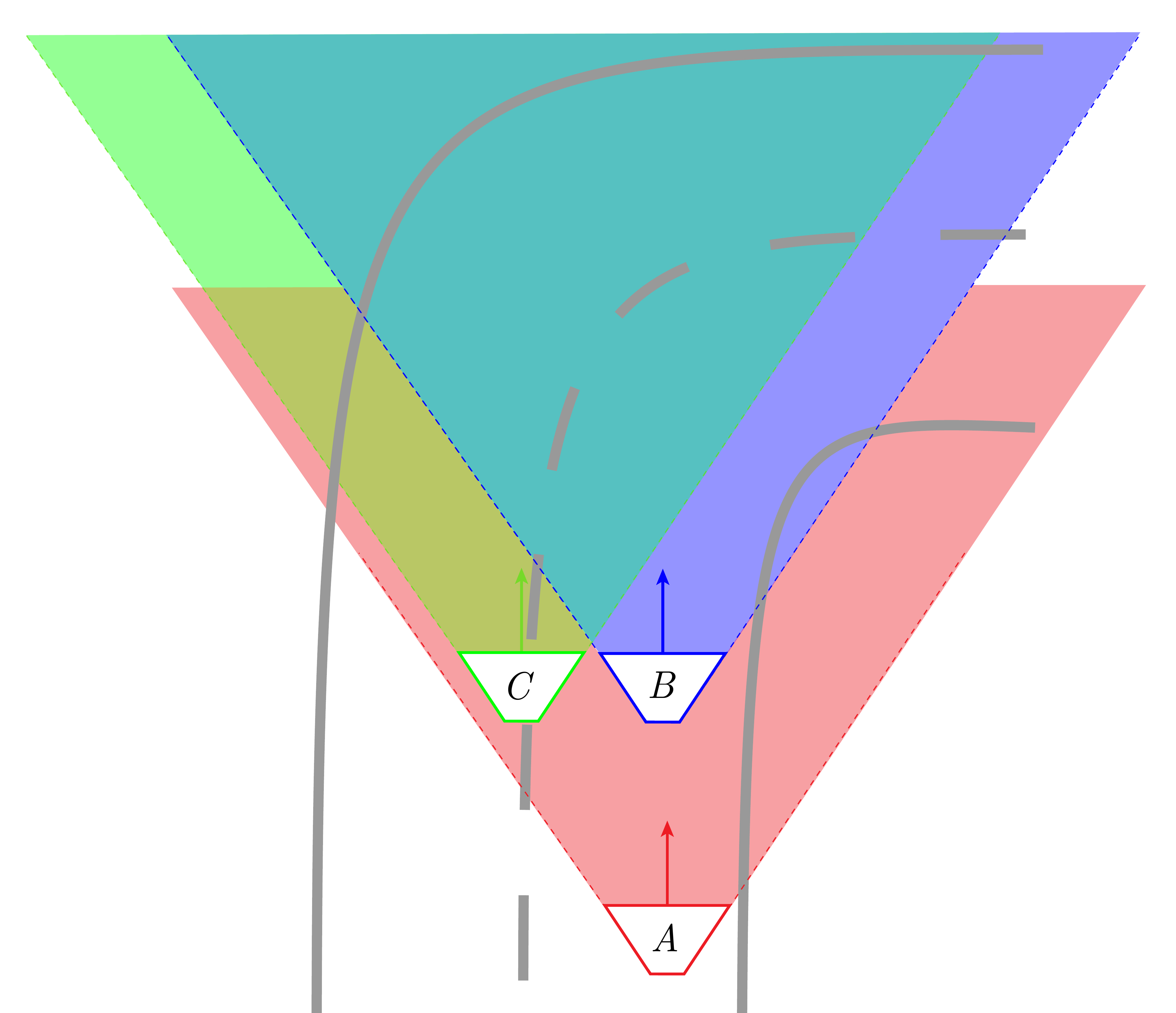}
    \end{subfigure}
    \begin{subfigure}{0.7\textwidth}
        \includegraphics[width=0.7\linewidth]{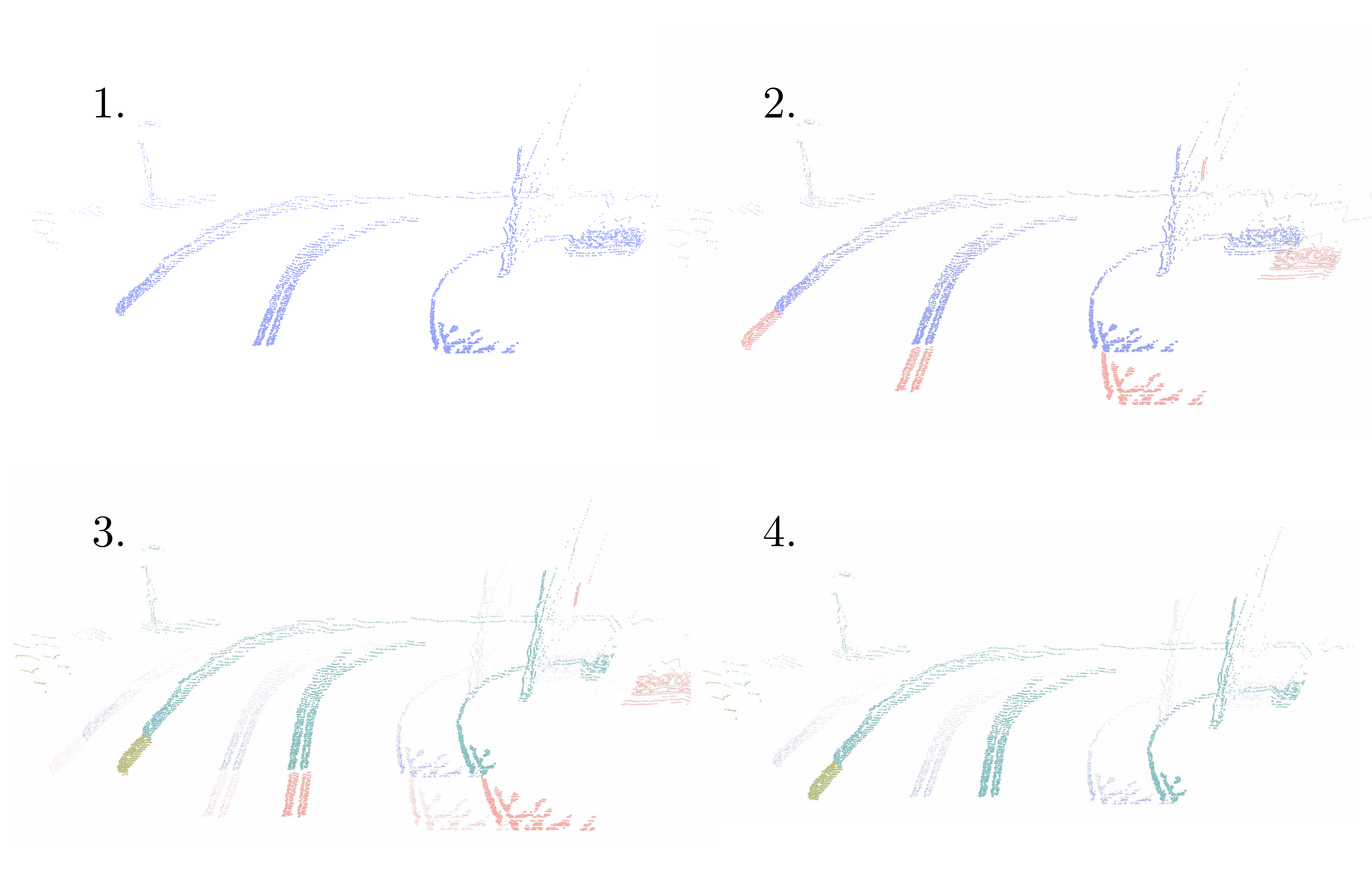}
    \end{subfigure}
    \caption{\textbf{Generation of a new camera frame. Top:} Visualization of the fields of view of different camera frames. Camera $C$'s field of view contains points not included in camera $B$'s field of view. These points are added by aligning a previous camera frame $A$. \textbf{Bottom: 1.)} Point cloud of camera $B$ (blue). \textbf{2.)} Aligning $A$'s point cloud (red) to add missing points. We choose $A$ to be far enough in the back to cover the field of view of $C$. \textbf{3.)} Changing the points' reference frame from $B$ to $C$. (The faded points represent the points with reference frame $B$.) \textbf{4.)} Finalizing the point cloud creation by removing all points outside the field of view of $C$.}

    \label{fig:point-cloud-generation}
\end{figure}

After obtaining the camera poses and point clouds for each camera frame, we finally have all prerequisites to generate additional trajectories needed to train the vehicle control model. Additional trajectories can be generated by laterally translating the point clouds from the base frame in the reference trajectory to a new frame position. Note that the reference trajectory's point cloud does not necessarily contain all the points encompassed by the field of view (FOV) of the new frame position. Those missing points are included from frames in the reference trajectory preceding the base frame. Conversely, points that do not fall within the field of view of the new camera frame position are discarded. The process is illustrated in Figure~\ref{fig:point-cloud-generation}, wherein Camera $B$ is the base frame in the reference trajectory, while Camera $A$ is a preceding frame also in the reference trajectory. We show how a point cloud at the new position represented by Camera $C$ can be synthesized. The process can be split into the following steps:

\begin{enumerate}
    \item Align the point cloud of a preceding frame in the reference trajectory with that of the base frame. 
    \item Shift the aligned point cloud from the base frame of reference to the new frame position 
    \item Remove all points outside the field of view of the new frame position.
\end{enumerate}
Next, we discuss these steps in detail.

\noindent{\textbf{Aligning Point Clouds}}\label{subsec:aligning}

Consider two camera frames \(A\) and \(B\). The \emph{global transforms} of the frames relative to the world frame are represented by \(T_A\) and \(T_B\). We have two local point clouds \(P_0\) and \(P_1\), which were captured from camera \(A\) and \(B\), respectively. We call \(P_0\) and \(P_1\) local point clouds, because the coordinates of their points are given relative to their respective reference frames \(A\) and \(B\).

Aligning point cloud \(P_0\) with point cloud \(P_1\) is synonymous to representing the coordinates of \(P_0\) relative to frame \(B\). The relative transform representing the rigid body motion from camera \(B\) to camera \(A\) is the same transform that changes the reference frame of the coordinates of point cloud \(P_0\) from \(A\) to \(B\).

Applying the transform \(T_A\) to \(P_0\) changes the reference frame of the coordinates of \(P_0\) from frame \(A\) to the world frame. It moves \(P_0\) to its global position. Let us call this global point cloud \(P_{0_W}\). Applying the inverse of \(T_B\) to \(P_{0_W}\) changes the reference frame for the coordinates of \(P_{0_W}\) from the world frame to frame \(B\). This new point cloud \(P_{0_B}\) is aligned with \(P_1\), as the coordinates of both point clouds are given relative to frame \(B\).

In summary, the transform we need to apply to \(P_0\) to align it to \(P_1\) is given by:

\begin{equation}
T_{BA} = T_{B}^{-1} T_A 
\end{equation}
Applying \(T_{BA}\) to \(P_0\) and concatenating the result with with \(P_1\) gives us a new point cloud \(P_{0,1}\), containing all the points of \(P_0\) and \(P_1\) with coodinates given relative to frame \(B\):

\begin{equation}
P_{0,1} = \begin{bmatrix} T_{BA} P_0 && P_1 \end{bmatrix}
\end{equation} 

If we choose \(A\) to be a camera frame earlier in the sequence than \(B\), then the scene represented by the combined point cloud \(P_{0,1}\) includes points from a wider range than \(P_1\). This allows us to generate a new camera frame by shifting and cropping the point cloud in the next step.\\

\noindent{\textbf{Shifting and Cropping Point Clouds}}\label{subsection:shifting_cropping}

To generate a new trajectory, we have to generate point clouds from the point of view of the camera frames in the new trajectory. This paper focuses on trajectories parallel to the base trajectory. Therefore, new camera frames are always shifted in lateral direction from the base camera frames. 

The field of view of a camera \(C\) shifted in lateral direction from camera \(B\) contains points not included in the field of view of camera \(B\). In the previous step, we therefore aligned previous point clouds to the point cloud of camera \(B\) to obtain a combined point cloud \(P_{0,1}\) including all points in the field of view of camera \(C\). The coordinates of \(P_{0,1}\) are given relative to frame \(B\) and its global transform is given by \(T_B\).

We want frame \(C\) to be shifted by some value \(x\) in lateral direction from frame \(B\). We will refer to frame \(B\) as the \emph{base frame}. The transform representing a lateral shift by value \(x\) is given by \(T_x\):

\begin{equation}
T_x = \begin{bmatrix}1 && 0 && 0 && x \\ 0 && 1 && 0 && 0 \\ 0 && 0 && 1 && 0 \\ 0 && 0 && 0 && 1\end{bmatrix}, x \in \mathbb{R}
\end{equation}

The relative transform of frame \(C\) to frame \(B\) is therefore given by \(T_x\). In general, \(T_x\) could be any rigid body motion but this paper focuses on trajectories parallel to the base trajectory and therefore on lateral shifts.
To obtain the global transform \(T_C\) of frame \(C\) relative to the world frame, we have to change the reference frame of $T_x$ from $B$ to the world frame. We do so by applying \(T_B\) to \(T_x\).

\begin{equation}\label{eq_tc}
T_C = T_B T_x
\end{equation}

\begin{equation}\label{eq_tx}
T_x = T_B^{-1} T_C
\end{equation}
Now we want to change the reference frame of \(P_{0,1}\) from frame \(B\) to frame \(C\). We can do so in two steps. First we change the reference frame of \(P_{0,1}\) from frame \(B\) to the world frame by applying \(T_B\). Then we can change the reference frame from the world frame to frame \(C\) by applying \(T_C^{-1}\). This two step transform is equal to applying \(T_x^{-1}\) directly to \(P_{0,1}\).

\begin{equation}
\begin{aligned}
P_{{0,1}_C} & = T_C^{-1}T_BP_{0,1} \\
            & = (T_B^{-1} T_C)^{-1} P_{0,1} \\
            & \stackrel{(\ref{eq_tx})}{=} T_x^{-1} P_{0,1}
\end{aligned}
\end{equation}

\(P_{{0,1}_C}\) is now at the position a local point cloud perspective of camera \(C\). But it also contains points outside the field of view of camera \(C\). To get a realistic point cloud, we need to discard the points whose projections lay outside the image plane of camera \(C\). 

To obtain the 2D coordinates of the points' projections, we multiply the intrinsic camera matrix (A) with the point cloud and divide each point by its depth to obtain its coordinates on the 2D image plane.

\begin{equation}\label{eq:2d}
    \begin{bmatrix}x_{im} \\ y_{im} \\ 1\end{bmatrix} = \frac{1}{Z} A\begin{bmatrix}X \\ Y \\ Z\end{bmatrix}
\end{equation}

Finally we drop each point from the point cloud for whose projection do not fulfill the following criteria: \[0 <= x_{im} < width \land 0 <= y_{im} < height\] where \(width\) and \(height\) denote the dimensions of the image plane.\\

\noindent{\textbf{Counteracting Imperfect Camera Poses}}\label{subsec:counteracting}

The larger the distance of the new frame \(C\) to its base frame \(B\), the more points in the field of view of camera \(C\) are not laying in the field of view of camera \(B\). Therefore the larger the distance, the more point clouds from camera frames earlier in the sequence need to be aligned to include their points in the generated point cloud of the new camera frame \(C\).

\begin{figure}
    \begin{subfigure}{0.5\textwidth}
        \centering
        \includegraphics[width=0.8\linewidth]{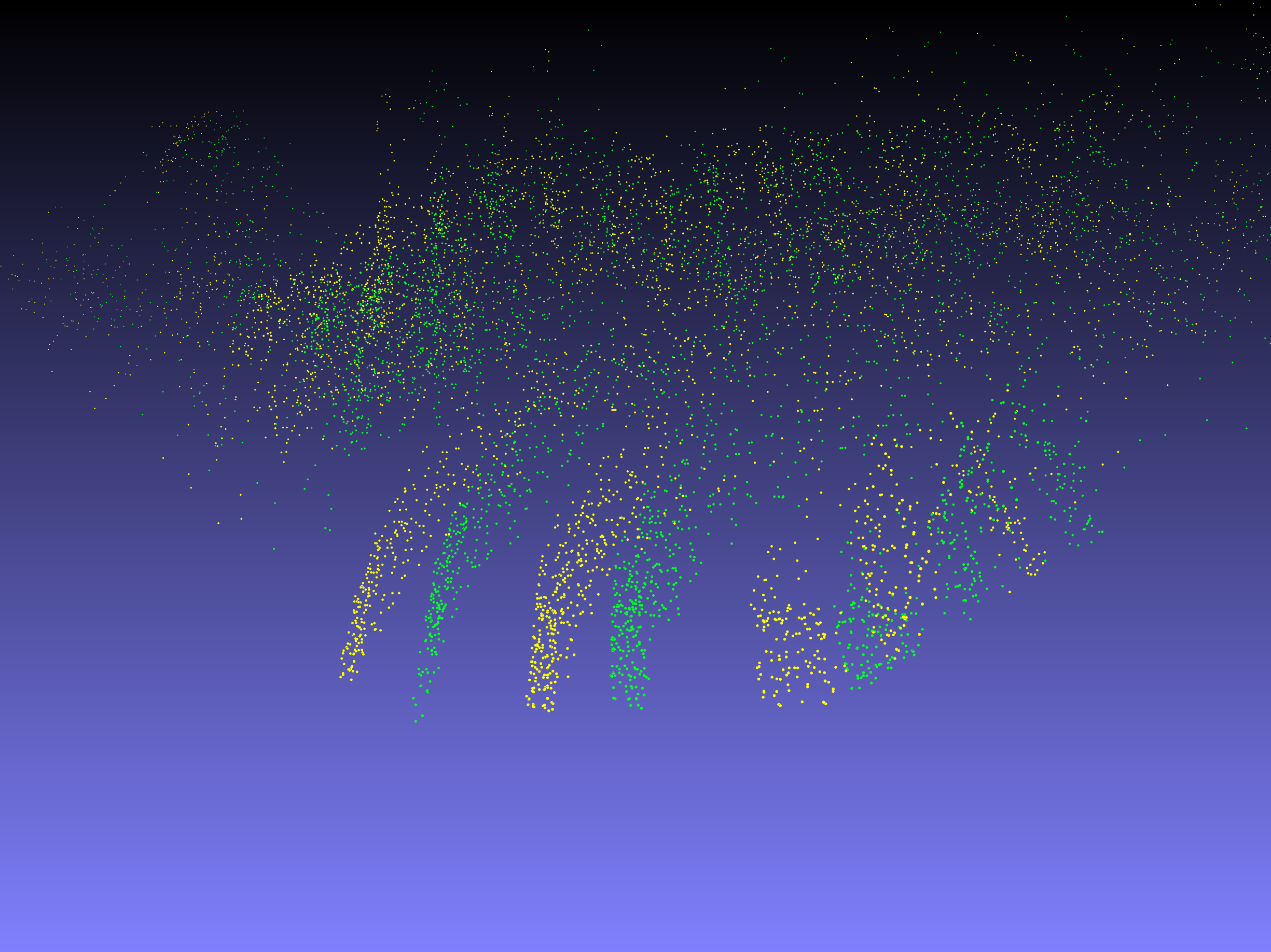}
        \caption{}
        \label{fig:fuzzy-pointcloud}
    \end{subfigure}
    \begin{subfigure}{0.5\textwidth}
        \centering
        \includegraphics[width=0.8\linewidth]{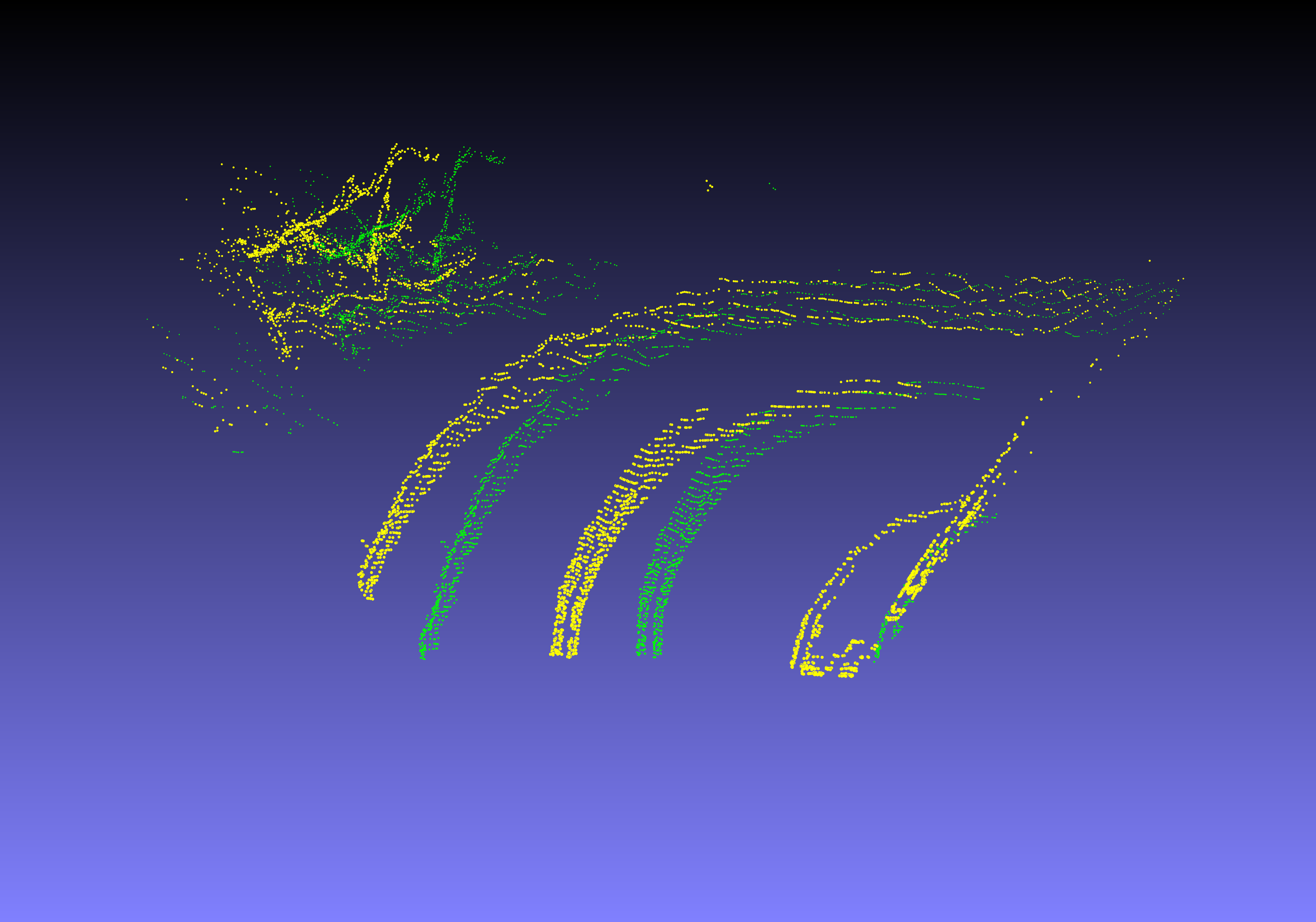}
        \caption{}
        \label{fig:crisp-pointcloud}
    \end{subfigure}
    \caption{\textbf{Counteracting Imperfect Camera Poses.} \textbf{a)} Multiple edge filtered point clouds are aligned to generate a new camera frame. Due to imperfect predicted camera poses, the resulting point cloud looks very fuzzy. \textbf{b)} By only adding points outside the field of view of the original camera and limiting the maximum distance, we can increases sharpness of the point clouds. Moreover, the maximum distance is limited to a threshold, as the depth prediction becomes less accurate with greater distance and closer points are more relevant for the prediction of the lateral offset of the next frame.}
    \label{fig:counteracting-imperfect-camera-poses}
\end{figure}

\noindent The camera poses predicted by visual odometry may deviate slightly from the true poses and the further we go back in the sequence, the more significant this deviation becomes. If we just align the previous point clouds and add all their points to the generated point cloud, the result looks quite fuzzy (see Figure~\ref{fig:fuzzy-pointcloud}).

\noindent To counteract this issue, we can remove all points from previous, aligned point clouds that lay within the field of view of our base frame. Consider camera frame \(A\), which is earlier in the camera sequence than camera frame \(B\). Let \(P_0\) be the point cloud captured from camera \(A\) and \(P_1\) the point cloud captured from camera \(B\). After aligning \(P_0\) to frame \(B\), we obtain the 2D coordinates of the projections of its points onto the image plane of camera \(B\) as described in Equation~\ref{eq:2d}. Then we drop each point from \(P_0\) for whose projection the following criteria \textbf{does} hold:  \[0 <= x_{im} < width \land 0 <= y_{im} < height\] where \(width\) and \(height\) denote the size of the image plane of $B$. We only keep the points whose projection lies outside the image plane of camera \(B\). This ensures that missing points are being added, but the existing point cloud is not blurred (see Figure~\ref{fig:crisp-pointcloud}).

\subsection{Label Generation}\label{subsection:label_generation}

We have already discussed the process of preparing the training data and generating additional trajectories. The next step is to generate the target labels for this data in order to be able to train a model for lateral vehicle control. Note that the camera is rigidly attached to the car, therefore the camera pose for any frame can also be used to determine the pose of the car at the corresponding timestep. For each car pose, we would like to find the appropriate steering angle to be executed such that a subsequent car pose is attained. We model the dynamics of front wheeled driven car using the bicycle model \cite{wang2001}. We assume the no-slip condition between the front and rear wheels \cite{rajesh2012}. This holds true when the car is moving straight or making turns at low/moderate speeds. Then the steering angle($\delta$) of the car is described by the equation \cite{pmlr-v130-khan21a}:
\begin{align}
    \delta = \tan^{-1}(\Delta x \cdot \alpha)
    \label{eq:final}
\end{align}
Where $\Delta$x is the lateral distance between 2 frames. This can be determined from the camera poses. The 2 frames are chosen such that they are approximately a fixed longitudinal distance apart. This is to cater for the car moving at variable speed or in case of dropped camera frames during data collection. $\alpha$ is a constant that can be calibrated at inference time depending on the car.

\subsection{Model Architecture}\label{subsection:pointnet}

\begin{figure}[h]
    \centering
    \includegraphics[width=0.5\textwidth]{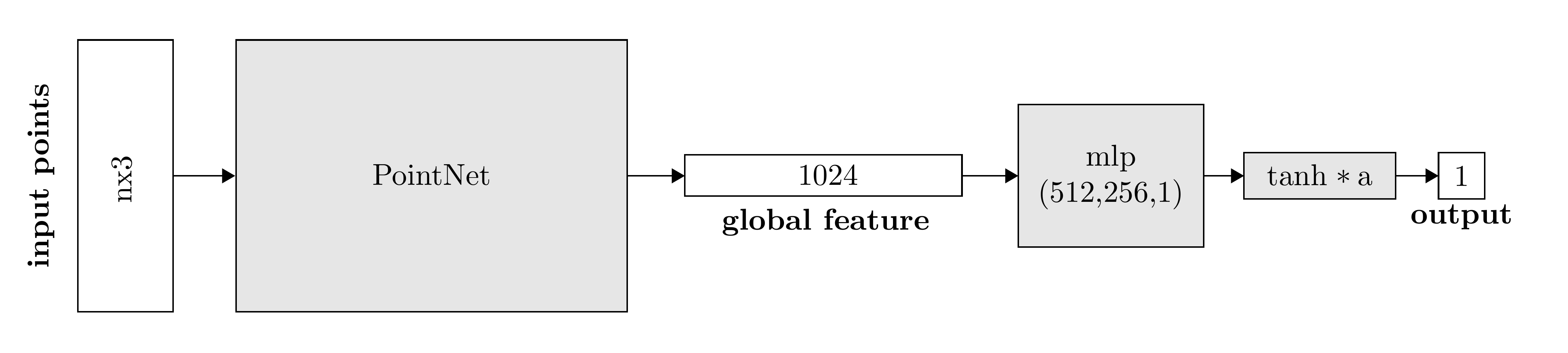}
    \caption{Shows the architecture of the model which takes a point cloud as input and predicts the lateral deviation ($\Delta$x) to a subsequent frame. \cite{qi2016pointnet} is used as the base architecture to produce a feature vector with length $1024$ from the point cloud with $n$ points. Meanwhile, "mlp" represents a multi-layer-perceptron, with the numbers in brackets representing layer sizes. "tanh" is a hyperbolic tangent function. It is scaled by a fixed value \emph{a} to allow predictions in the range of \emph{-a} to \emph{a}.}
    \label{fig:model-architecture}
\end{figure}

Using the generated point clouds and labels, a deep learning model is trained to predict the lateral deviation ($\Delta$x) between the current frame and a subsequent frame.  The model can be considered as comprising of 2 main components. The first takes raw point clouds of the current frame as input and produces a global feature vector to furnish a latent representation of the scene  as seen by the ego-vehicle. We adapt the PointNet architecture \cite{qi2016pointnet} prior to the classification head for this. The next component consists of multi-layer perceptions which map this global features vector to furnish the output.   A hyperbolic tangent function forms the last layer. It is scaled by an arbitrary fixed value \emph{a} to allow predictions in the range of \emph{-a} and \emph{a}. The loss function chosen to train the model is \emph{mean squared error} of the lateral deviation ($\Delta$x) between the 2 frames and that predicted by the model. Figure~\ref{fig:model-architecture} shows a visualization of the model architecture.

\section{Experiments}\label{chapter:experiments}

We use the CARLA (CAR Learning to Act) \cite{dosovitskiy2017carla} simulator (stable version 0.8.2) for our experiments. It allows for an online evaluation to assess the true driving quality. This is as opposed to offline evaluation where 2 models with the same offline metrics can have drastically different driving performance \cite{codevilla2018offline}.

\begin{figure}[h]
    \centering
    \includegraphics[width=0.5\textwidth]{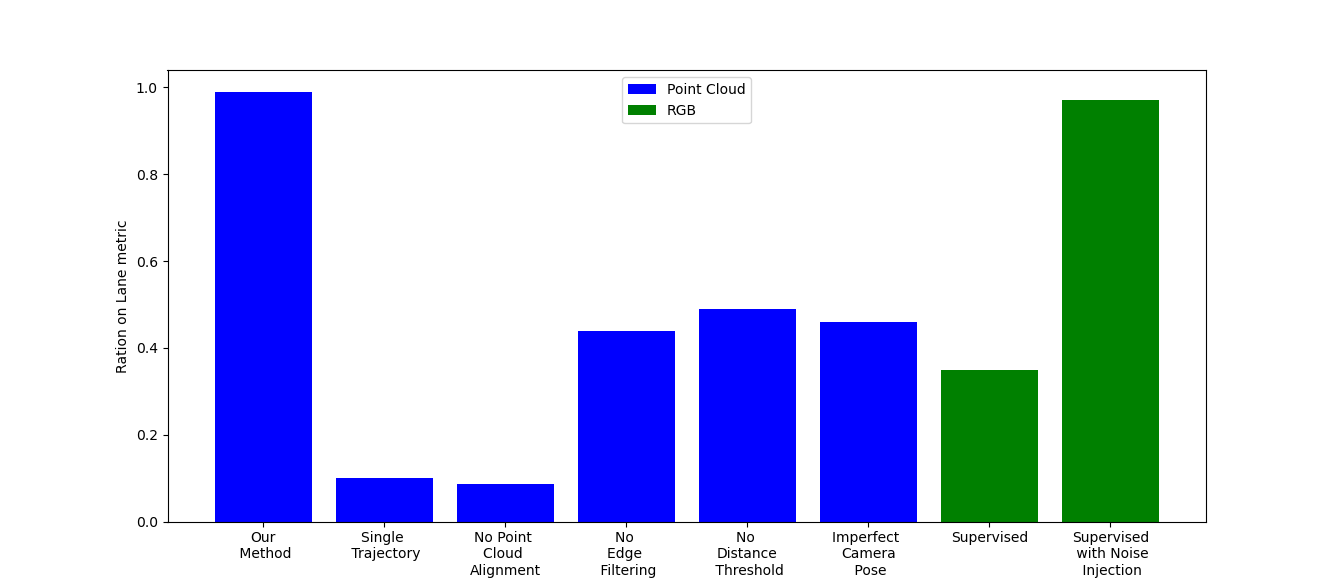}
    \caption{Average ratio of frames the ego-vehicle remains within its driving lane for different model configurations across different starting positions unseen during training. Higher value is better. Note that our method is better than the supervised and at par with the supervised approach with noise injection \cite{codevilla2017endtoend}. However, in contrast to this noise injection strategy, our method does not run the risk of traffic violations during the data collection phase. This is further explained in Section \ref{chapter:discussion}}
    \label{fig:results}
\end{figure}

The unit of measurement we use for online evaluation is the \emph{ratio on lane} metric adapted from \cite{8968451}. It gives the ratio of frames the ego-vehicle remains within its own driving lane to the total number of frames. The ego-vehicle is considered driving within its own lane if no part of the bounding box of the car is on the other lane or off the road and the car is not stopped due to a collision with other traffic. \\
\textbf{\noindent{Data Collection: }}\\ Image data is collected at a fixed frame rate by traversing the ego-vehicle in auto-pilot mode in Town 01 of the CARLA simulator. Images of size 640 x 192 (90\textdegree FOV) are used to train the depth estimation network using \cite{monodepth2}. Meanwhile,  \cite{schoenberger2016sfm} determines the camera poses. The depth is used to generate a 3D point cloud which in turn produce additional shifted point clouds at off course trajectories. We generate ten additional point cloud trajectories at uniform lateral distances in the range of [-2,2] meters from the single reference trajectory. The missing points at shifted point clouds can be compensated for by aligning preceding point clouds using the camera poses. Only the missing points from the preceding frames are added. This is because imperfect camera poses may result in duplication of objects, thereby confusing the model. The camera poses additionally allow to determine the target labels to train the network. The number of points input to the network are fixed to 4096. This is done by filtering the point cloud to retain only the high level edge features, while points beyond 20m are discarded.\\  
\textbf{\noindent{Quantitative Evaluation: }}\\
The evaluation at inference time is done at different starting positions in the town that are unseen during training. Each episode is executed for 135 frames at fixed throttle and the mean ratio on lane metric is reported across all unseen episodes. Results of our method are reported in Figure \ref{fig:results}.

Moreover, comparison with different point cloud configurations is also done to see the impact of the various components of our framework on the overall driving performance. Configurations explored are: training with a single trajectory rather than with multiple generated trajectories, aligning preceding point clouds to generate a new trajectory versus only shifting, limiting the maximum depth distance, using an edge filter to filter the point cloud and dealing with imperfect camera poses. We additionally compare with supervised RGB model baselines. Therein, we demonstrate that the performance of our framework is superior to the supervised model and at par with the supervised model trained with noise injection \cite{codevilla2017endtoend} which we explain in Section \ref{chapter:discussion}.
\section{Discussion}\label{chapter:discussion}
In this section, an explanation of the various configurations given in Figure \ref{fig:results} are described. The consequence of these configurations on the online driving performance are also discussed. 

\begin{figure}[h]
    \centering
    \includegraphics[width=0.51\textwidth]{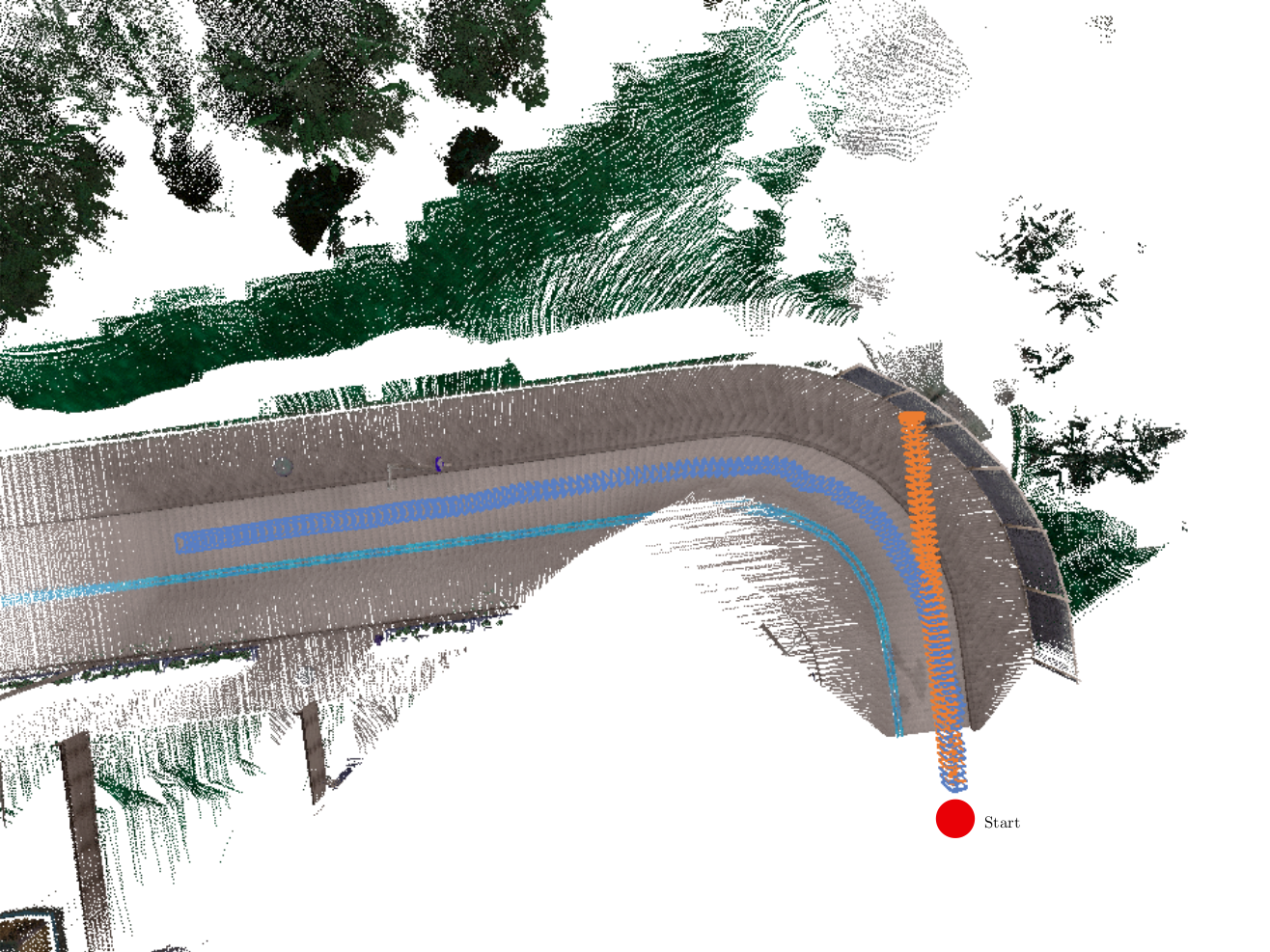}
    \caption{\textbf{Path visualization}. The red dot marks the starting position. \emph{Out method} (blue) was trained using multiple generated trajectories and is able to follow the curve. Meanwhile, the \emph{Single Trajectory} (orange) model was trained using only the single reference trajectory and drives straight into the barrier.}
    \label{fig:bev-m-vs-s}
\end{figure}

\noindent{\textbf{Single Trajectory v. Multiple Generated Trajectories: }}\\The core of our proposed method revolves around the capability of generating additional trajectories from a single reference trajectory, as described in Section~\ref{subsection:data_generation}. We therefore evaluate the impact of this by comparing with the single trajectory model trained only on the reference trajectory. It can be seen that the ratio on lane metric for the single trajectory is far inferior in comparison to our multi trajectory model. One plausible explanation is that when the model deviates off course, it cannot make the appropriate correction, since the reference trajectory data does not capture such scenarios during training. Figure \ref{fig:bev-m-vs-s} shows a birds eye view (BEV) visualization of the models’ paths at one of the starting positions in the training set. While
our model is able to follow the road, single trajectory model crashed straight into the barrier.\\

\noindent{\textbf{Aligning Point Clouds when Generating new Trajectories: }}\\Section~\ref{subsection:data_generation} described the process of generating point clouds in new trajectories by aligning point clouds from a preceding frame and then shifting.  To examine the necessity of aligning previous point clouds, we train another model whose point clouds were generated without aligning previous point clouds but instead only shifting and cropping the reference trajectory. Shifting the point cloud will yield empty regions that cannot be captured within the FOV of the source camera in the reference trajectory. This is particularly true of trajectories that are generated at farther distances away from the reference. It is further exacerbated for source images captured when the car is executing turns. As can be observed, the performance of such a model trained with partially observable point clouds drops dramatically.

\noindent{\textbf{Edge Filtered v. Full Point Cloud: }}\\The authors of \cite{8968451} alluded to high level features such as lane markings, sidewalk/road intersections, barriers etc. being important for the vehicle control model to hold its driving lane. However, in uncolored point clouds those features are clearly less visible than in RGB images. This is why it is important to make them more prominent to the model and increase the density of relevant information in the point cloud. This is done by filtering seemingly irrelevant points from the point cloud by applying an edge filter as described in Section~\ref{subsection:monodepth}. To evaluate the benefit of edge filtering, another model was trained on the full, unfiltered point clouds. The ratio on lane performance of this model drops, thereby advocating in favor of our edge filtering approach. 

\noindent{\textbf{Limiting the Distance: }}\\Section~\ref{subsection:monodepth} also mentioned limiting the distance of points to the camera as part of an approach to increase concentration of points in relevant areas closer to the camera. If a model is trained  without a maximum point distance, its performance drops. One tenable explanation for this is that at larger distances, the depth prediction is less certain. This imperfect depth could result in object duplication at farther distances when aligning point clouds. This can be observed in the top right region of the 2\textsuperscript{nd} point cloud in Figure \ref{fig:point-cloud-generation}. Hence, when such anomalies at farther depths are removed from the point cloud by limiting the maximum distance, the performance of the model is enhanced. Moreover, points in the immediate vicinity of the ego-vehicle are more important for executing the appropriate steering command rather than the points farther away that have imprecise depths.

\noindent{\textbf{Dealing with Imperfect Camera Poses}}
When aligning previous point clouds as part of the process of generating a point cloud for a new trajectory, imperfect camera pose estimations can cause blurriness if objects in the 3D space do not align. Subsection~\ref{subsec:counteracting} described an approach of counteracting this blurriness by only adding missing points outside the base camera frame's field of view instead of all points. To evaluate the effectiveness of this approach, we trained another model wherein all points are used for training. As can be seen the performance of such a model drops.

\begin{figure}[h]
    \centering
    \includegraphics[width=0.5\textwidth]{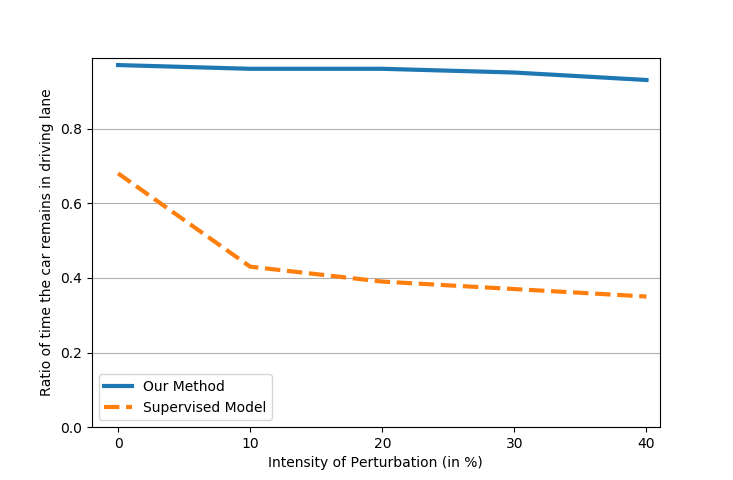}
    \caption{\textbf{Avg. ratio on lane for different levels of perturbation on the training trajectories.} The performance of both the supervised RGB model (orange) and Our method (blue) decreases with higher intensity of perturbations, but the descent is more dramatic for the supervised model.}
    \label{fig:pc-vs-rgb-noise-avg}
\end{figure}

\noindent{\textbf{Supervised RGB model: }}\\We additionally train a supervised convolutional model to compare with our framework using point clouds. This new model takes in an RGB image as input and directly predicts the appropriate steering command for vehicle control. In contrast to our method this RGB model is supervised and uses the ground truth steering commands as labels during training. The performance of this model is much lower than our method.  This is despite the fact that our model was not trained with any ground truth steering labels. Therefore, the superior performance of our method can be attributed to the ability to generate additional off trajectory training data from a single traversal of the ego-vehilce causing it to be more robust.\\
 \emph{Affect of Perturbations: } To further compare the robustness of our method with the supervised RGB model, we add perturbation of varying degree into the steering command predicted by the model at each time step. The amount of perturbation is sampled from a uniform random distribution. Figure \ref{fig:pc-vs-rgb-noise-avg} shows that as magnitude of perturbation is enhanced, our model maintains a fairly consistent performance. In contrast, the supervised RGB model has a steep performance drop. This is because when perturbations are introduced the ego-vehilce may drive off-course for which the supervised model is not capable of counteracting such situations.  On the other hand, our model is accustomed to handling such situations as it was fed additional off trajectory point clouds during training. \\

\begin{figure}[h]
    \centering
    \includegraphics[width=0.5\textwidth]{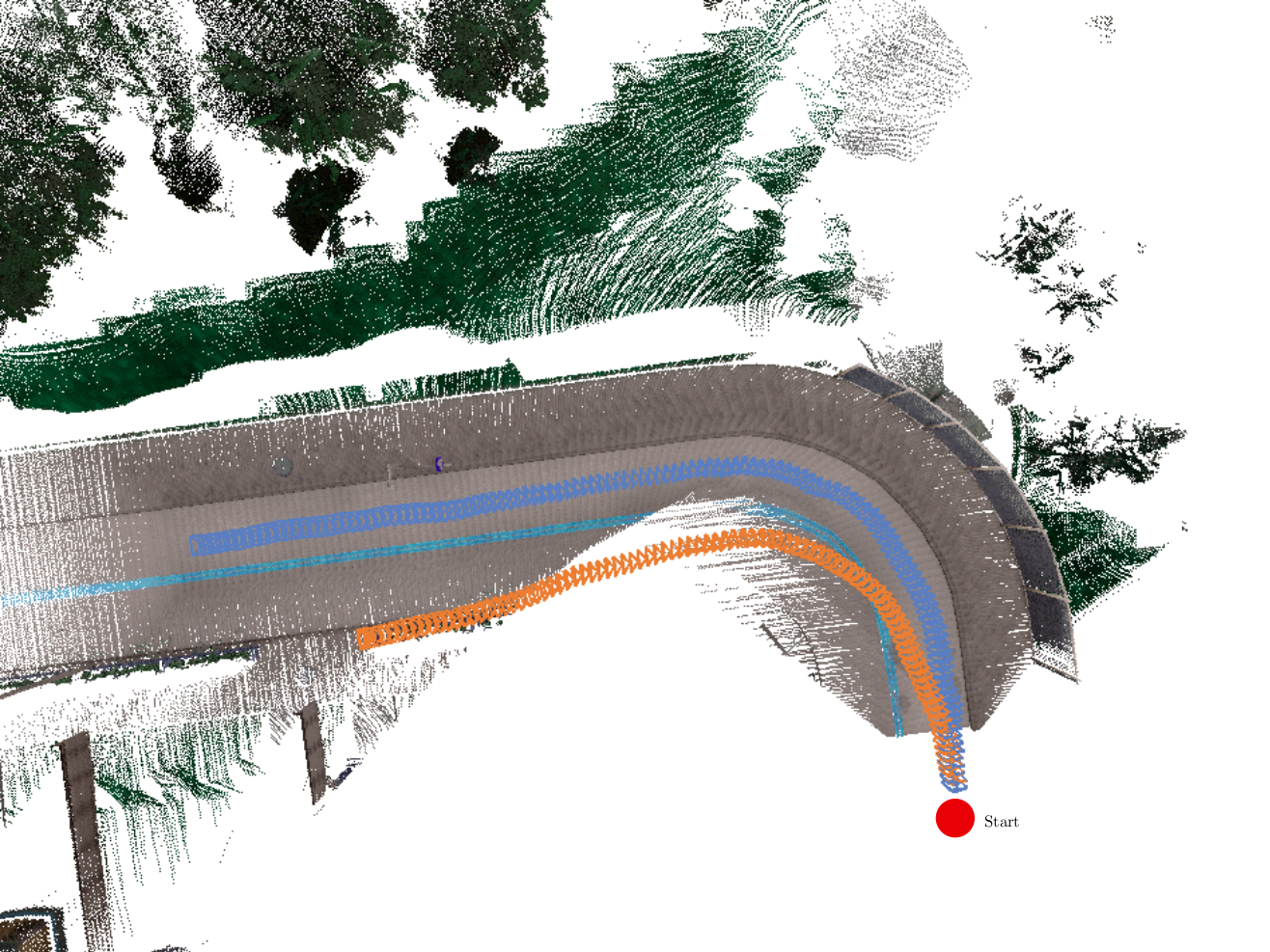}
    \caption{\textbf{Path visualization on a trajectory in the training set with noise level of 10\%.} The blue path belongs to the model trained with \emph{Our Method}, while the orange path represents the \emph{Supervised RGB} model. The red dot marks the starting position. \emph{Our Method} is able to keep the lane despite added noise, while \emph{Supervised RGB} leaves the lane shortly after the start and is not able to recover. }
    \label{fig:pc-vs-rgb-bev-noise}
\end{figure}

Figure \ref{fig:pc-vs-rgb-bev-noise} shows a BEV visualization of the paths of Our (blue) and the supervised RGB (orange) models when noise is added in one of the training trajectories. Despite the perturbations, our model is able to keep the
car on its lane. Meanwhile, the supervised RGB model looses track shortly after the start,
is not able to correct its path and eventually crashes.

\noindent{\textbf{Supervised model with noise injection:}}\\ We additionally compare our approach with the supervised RGB model proposed by \cite{codevilla2017endtoend}. It is similar to the supervised model described above except that during data collection noise is injected into the steering command. This causes the vehicle to diverge from its normal path. The corrective steering maneuver taken by the expert driver to limit this divergence and bring the vehicle back on course is recorded. Performance of such a model is comparable to our approach. This is because the data used in training such a model contains off-trajectory images and corresponding labels. However, executing such a strategy for data collection with traffic participants may be extremely dangerous and may even involve violation of traffic rules. Moreover, it involves having an expert driver capable of handling dramatic maneuvers resulting from this noise injection. Most importantly, since our approach does not require any noise injection, it does not run the risk of violating traffic rules during data collection.\\

\section{Conclusion}\label{chapter:conclusion}
In this paper, we put forth a framework to train a point cloud based deep learning model on the task of lateral control of an autonomous vehicle. The model is capable of learning a robust driving policy from merely an unlabeled sequence of RGB images, captured with a single front facing camera on a single reference trajectory. The efficacy of our approach comes from the capability of generating additional data from the same sequence. The additional data appears as if it emerges from an off course trajectory. This  counteracts the limitations of imitation learning which suffer from insufficient robustness in the real world due to a lack of failure cases in the training data. Online experiments on the  driving simulator showed that its performance is superior to a supervised baseline CNN trained on the same initial data set. Given no labels are required for the training data, the approach is scalable to large quantities of data. This makes it a more robust alternative to current supervised end-to-end lateral control methods.\\

\noindent{\textbf{Acknowledgement:}}\\
This work was supported by the Munich Center for Machine Learning.

\bibliographystyle{IEEEtran}
\bibliography{main} 
\end{document}